\documentclass[journal]{IEEEtran}
\usepackage{amsfonts}
\usepackage{amsmath}
\usepackage{txfonts}
\usepackage{amssymb}
\usepackage[compress]{cite}
\usepackage{graphicx}
\usepackage{algorithm}
\usepackage{algpseudocode}
\usepackage{subfigure}
\usepackage{float}
\usepackage{subfig}
\usepackage{array}
\usepackage{booktabs}
\usepackage{multirow}
\usepackage{epsfig}

\begin{document}
\title{Emerging Applications of Reversible Data Hiding}

\author{
\IEEEauthorblockN{${\mathrm{Dongdong~Hou}}^{1}$, ${\mathrm{Weiming~Zhang}}^{2}$, ${\mathrm{Jiayang~Liu}}^3$, ${\mathrm{Siyan~Zhou}}^4$, $\mathrm{Dongdong~Chen^5}$, ${\mathrm{Nenghai~Yu}}^{6}$}\\
\IEEEauthorblockA{$\mathrm{^{12356}School~ of ~Information~ Science ~and ~Technology, University ~of ~Science~
and ~Technology~ of~ China, ~Hefei,~ China}$\\
$\mathrm{^{4}School~ of~ Computer~ Science~ and~ Software~ Engineering,~ East~ China~ Normal~ University,~ Shanghai,~ China}$\\
\{
Email: $^{1}\mathrm{houdd@mail.ustc.edu.cn}$, $^{2}\mathrm{zhangwm@ustc.edu.cn}$, $^{3}\mathrm{ljyljy@mail.ustc.edu.cn}$, $^{4}\mathrm{51184506084@stu.ecnu.edu.cn}$, $^{5}\mathrm{cd722522@mail.ustc.edu.cn}$, $^{6}\mathrm{ynh@ustc.edu.cn}$.\}}
}

\maketitle
\begin{abstract}
   Reversible data hiding (RDH) is one special type of information hiding, by which the host sequence as well as the embedded data can be both restored from the marked sequence without loss. Beside media annotation and
integrity authentication, recently some scholars begin to  apply RDH in many other fields innovatively. In this paper, we summarize these emerging   applications, including steganography, adversarial example,  visual transformation, image processing, and give out the general frameworks to make these operations reversible. As far as we are concerned, this is the first paper to  summarize the extended applications of RDH.
\end{abstract}
%%%%%%%%% BODY TEXT

%-------------------------------------------------------------------------
\section{Introduction}

\begin{figure}[h!]
\centering
\includegraphics[width=0.4\textwidth]{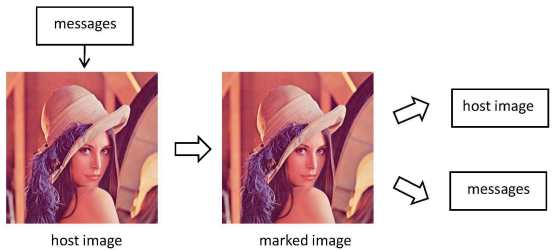}
\begin{center}
\caption{\small{Reversible data hiding.}}\label{RDH}
\end{center}
\end{figure}

Data hiding embeds messages into digital multimedia such as image, audio, video through an imperceptible way, which is mainly used for  copyright protection, integrity authentication, covert communication. Some special signals such as medical imagery, military
imagery and law forensics are so precious that cannot be damaged. To protect these signals,
reversible data hiding (RDH) \cite{RDH} is developed. Taking image as example (see Fig. \ref{RDH}), by RDH after embedding messages into the host image the generated marked image is visually invariant, and at the same time we can losslessly restore host image after extracting the embedded messages.

RDH algorithms are well established,  even the schemes \cite{RCC1,RCC2} approaching theoretical optimum have also appeared. As for RDH, the first step is to generate a host sequence with
the small entropy such as prediction errors (PEs) \cite{grayRDH1,grayRDH2,grayRDH3,colorRDH1,colorRDH2}, and then the users reversibly embed
messages into the host sequence by modifying its histogram \cite{HS,DE,RCC1}.

RDH is mainly used for media annotation and
integrity authentication, but its application is now extended by scholars. With RDH we can restore both embedded messages and host image, this make the host image  like storage disk which can be erasable. However, marked image generated from RDH is hard to resist detection. If we endow RDH with  undetectability,  and then such RDH algorithms called reversible steganography can be applied for convert storage \cite{keniyinxie1,keniyinxie2,keniyinxie3}. Besides convert storage, we can also regard RDH as one tool to do many reversible image operations. To be detailed, after operating  image to the desired target, we can explore the auxiliary parameters for restoring the original image from the target image, and then reversibly embed the parameters into the target image to get the reversible operated image. At the receiver's side, we extract these  auxiliary parameters and the target image from the reversible operated image, and further restore the original image from the target image with the extracted parameters.

%Apart from the applications of privacy protection,Image processing is popular, while almost all of them are not reversible. To solve it, we .

  In the following contents, we will  give out the general frameworks to do reversible steganography, reversible adversarial example, reversible  visual transformation and reversible image processing respectively.

\section{Emerging applications}\label{subsec2}

\subsection{Reversible steganography}\label{subsec2.1}
 The  popular method for privacy protection is encryption. But the messy codes of ciphertext with
special form are easy to cause the attention of attackers. Therefore, covert storage hiding the existence of data has been widespread concerned. It is clear that covert storage requires two properties at the same time, i.e., undetectability and reversibility.

Steganography is a secure tool designed for covert communication, and the most successful steganographic algorithms \cite{CMD,hill} are
devoted to embedding messages while minimizing the total distortion, which can achieve the strong undetectability under the advanced steganalysis \cite{spam,rich}.
 However, steganography will destroy host image irreversibly.   Different from covert communication, as for covert storage the image here is used as a special kind of storage medium that needs to be erasable like a disk. After deleting the stored data, the storage medium must be restored to its original state. To make the image erasable so that the storage space can be used repeatedly. Besides the undetectability, reversibility is also desired. From this point, RDH is suitable for covert
storage, by which the cover image can be losslessly restored after the message being extracted. But traditional RDH algorithms are not secure  under steganalysis.

The high undetectability of steganography is mainly achieved by modifying the complex regions of images. However,
 most of RDH algorithms give the priority of modifications to pixels in smooth regions due to that  pixels in
smooth areas can be  predicted more accurately. That's the reason why
traditional RDH cannot resist steganalysis. Recently, Hong \emph{et
al}. \cite{keniyinxie1} give out the first RDH algorithm which
has much higher undetectability than traditional ones. The  undetectability is mainly achieved by embedding  messages into
PEs with small absolute values, but PEs in complex regions are preferentially modified through a sorting technique. Then Zhang \emph{et al}. \cite{keniyinxie2} improve Hong \emph{et al}.'s method by giving the
priority to PEs with the larger absolute values for accommodating messages.

To endow RDH with the undetectability, we should define inconsistent distortion metrics for each pixel according to its local variance. Generally speaking,  the distortion metric in smooth region should be defined higher than that in noisy region. For the convenience of handling,  inconsistent distortion metrics are quantized as multi-distortion metrics. Assume that these inconsistent distortion metrics are clustered into $K$ ($K\leq N$, $N$ is the  volume of host elements) classes. Accordingly, the host  sequence ${\rm {\bf X}}$ is segmented into $K$ sub-sequences denoted as ${\rm {\bf x}_{i}}=(x_{i, 1}, x_{i, 2}, \cdots, x_{i, N_i})$, where $N_i$ is the volume of ${\rm {\bf x}_{i}}$. The elements in ${\rm {\bf x}_{i}}$ will share the same distortion metric $d_i(x,y)$, where $1 \leq i \leq K$.
Then after giving the embedding rate $R$, the rate-distortion problem of RDH under multi-distortion metrics is  formulated as
\begin{equation}\label{RDB}
\begin{array}{ll}
\mbox{minimize} &
\frac{\sum_{i=1}^K N_i\sum_{x=0}^{m-1}\sum_{y=0}^{n-1}P_{X_i}(x)P_{Y_i|X_i}(y|x)d_i(x,y)}{N} \\
\mbox{subject to} &  \begin{array}{lll}
% H(Y_i)=R_i+H(X_i), & 1\leq i\leq K \\
 \frac{\sum_{i=1}^K N_i\times H(Y_i) }{N}=R+\frac{\sum_{i=1}^K N_i\times H(X_i) }{N}\\
\end{array}  %
\end{array}.
\end{equation}

\begin{figure}[h!]
\centering
\includegraphics[width=0.5\textwidth]{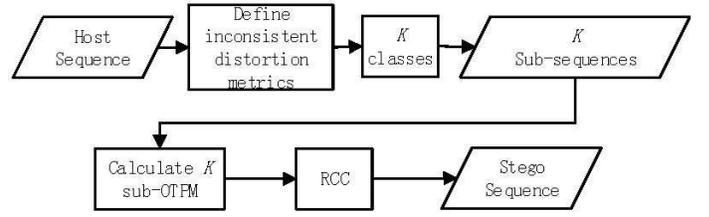}
\begin{center}
\caption{\small{Reversible steganography.}}\label{stegflow}
\end{center}
\end{figure}

 To minimize the average distortion between the host sequence ${\rm {\bf X}}$ and the generated marked sequence ${\rm {\bf Y}}$, each   sub-OTPM $P_{Y_i|X_i}(y|x)$ is desired for $i= 1, 2, ..., K$, by which we can optimally modify the $K$ sub-sequences and embed the corresponding messages into each sub-host-sequence. The unified framework for estimating the corresponding  $K$ sub-OTPMs $P_{Y_i|X_i}(y|x)$ is  presented in \cite{keniyinxie3}. After getting these sub-OTPMs, we perform recursive code construction (RCC) to finish the message embedding. Reversible steganography \cite{keniyinxie1, keniyinxie2, keniyinxie3} has the reversibility as traditional RDH and
the ability of undetectability as traditional steganography, which will be rather valuable in the applications of covert storage.
 Based on the above, one  framework of reversible steganography minimizing the distortion is given  as Fig. \ref{stegflow}.

%Taking images as example, the host image will be modified to create one adversarial example. The existing algorithms will modify host image irreversibly.
\subsection{Reversible adversarial example}\label{subsec2.2}
As for one deep neural network (DNN), its parameters are optimized  by reducing the loss gradually on the training set. After optimizing these parameters, DNN can be successful applied in  image classification, speech recognition, and so on. However, as for the trained network, one input with the carefully selected small perturbation perhaps will get a completely different result. Such input to fool the network is called an adversarial example. One can perturb an image to misclassify network while keeping the image quality to be imperceptible for human eyes. Even worse, an attacker can use the trained classifier to generate adversarial example, and then use it to fool another model.

The users create adversarial example to fool DNNs, but the created adversarial example must be controlled the users themselves. That is to say, the creators must be able to restore the adversarial example and can not let adversarial example to fool their own networks.  Usually, the image to generate adversarial example is the sensitive and important one such as military imagery, which  can not be damaged with  loss. %Besides DNNs, there are still many other recognition systems,
 After modification the adversarial example  can be deemed as a noisy image, although the distortion sometimes may be not sensitive to human eyes. However, the further  processing on adversarial example must will be interfered. For example,  the adversarial example must
will affect the feature extraction and thus result in the decrease
of processing accuracy. As for some important applications such
as military system, police system, and so on, the slight decrease of  accuracy perhaps will result in serious consequences. Therefore, reversible adversarial example is desired.

One simple way for creating adversarial example proposed by Goodfellow \emph{et al} \cite{gradAE}  is adding the perturbation on host image, and the added perturbation is the direction in image space which yields the highest increase
of the linearized cost. The hyper-parameter $\varepsilon$ is applied to limit the distance between the adversarial image $\mathbf{\emph{X}}_{adv}$ and the
host image $\mathbf{\emph{X}}$. Specifically,
\begin{equation}\label{RVThou}
\mathbf{\emph{X}}_{adv} = \mathbf{\emph{X}} + \epsilon\cdot sign(\nabla_{\mathbf{\emph{X}}}J(\mathbf{\emph{X}}; y_{true})),
\end{equation}
where $y_{true}$ is the target fooling class.

\begin{figure}[h!]
\centering
\includegraphics[width=0.45\textwidth]{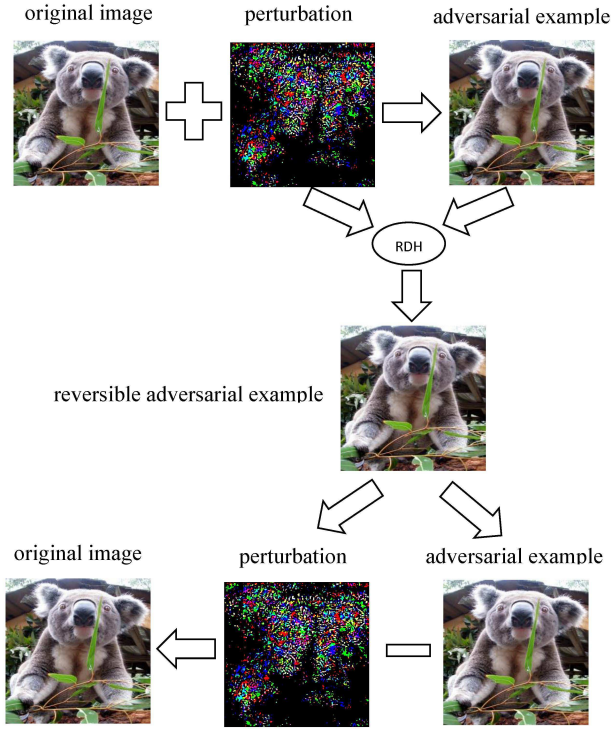}
\begin{center}
\caption{\small{Reversible adversarial example.}}\label{RAE}
\end{center}
\end{figure}

In \cite{RAE}, we firstly present the concept of reversible adversarial example, whose framework is  shown as  Fig. \ref{RAE}.   We reversibly embed the perturbation into  adversarial example to get  reversible adversarial example. To restore the original image, we first restore  adversarial example after extracting  the perturbation from  reversible
adversarial example, and further subtract  perturbation from adversarial example to return the original image.

\subsection{Reversible visual transformation}\label{subsec2.3}

Traditional secure data hiding algorithms \cite{CMD,hill} are effective for embedding a small part of messages into a
large cover, such as an image. But for image transmission and storage, the image itself is the secret file to be protected. Therefore, we need large capacity data hiding method to hide image.  Reversible visual transformation is usually used for image protection, which reversibly transforms a secret image to a freely-selected target image with the same size and gets a camouflage image similar to the target image.

Lai \emph{et al}. \cite{lai} propose the first work about visual transformation, which generates transformed image by replacing each target block with one corresponding similar secret block, and then record the location mapping accounting for the main auxiliary information with RDH. By this method, the target image must be similar to the secret image. What is more, the camouflage image is in the poor visual quality.

\begin{figure}[h!]
\centering
\includegraphics[width=0.45\textwidth]{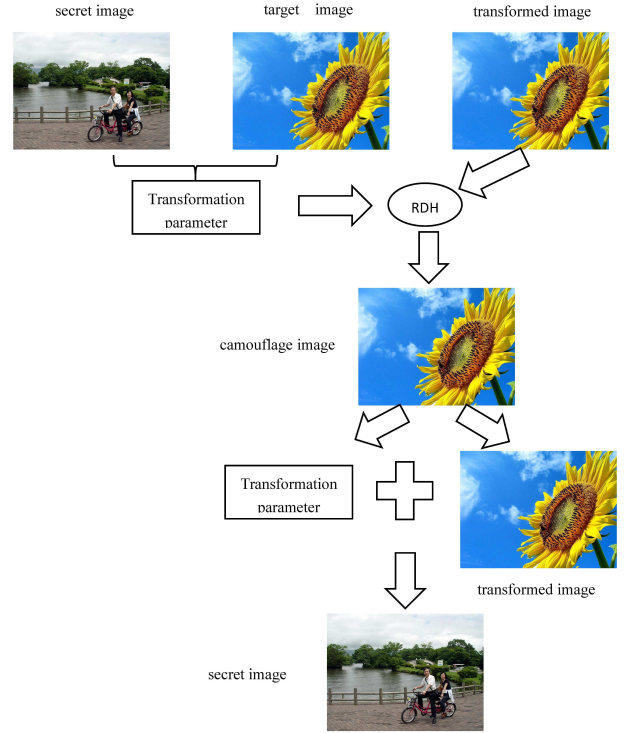}
\begin{center}
\caption{\small{Reversible visual transformation.}}\label{RVTflow}
\end{center}
\end{figure}
Generally speaking,  as long as the mean and the standard deviation  of each secret block are adjusted to be similar  with those of the corresponding target block, the secret image will be masqueraded as the target image visually. Lee \emph{et al}'s method \cite{lee} can  transform a secret image to a freely selected
target image via  color transformation \cite{colorRVT}, which greatly improves the visual quality of camouflage
image. However, by Lee \emph{et al}.'s
method  the secret image cannot be losslessly reconstructed due
to that the adopted color transformation \cite{colorRVT} is not reversible.

To make the transformation reversible,  a novel reversible visual transformation (RVT) scheme is presented by using shift transformation \cite{hou1}. Before shift transformation, a
non-uniform clustering algorithm is utilized to match secret blocks and
target blocks, which largely reduces the amount of accessorial information
 for recording indexes of secret blocks. To further reduce the amount of accessorial information, the correlations among  three color channels and  inside each color channel are explored \cite{hou2}. Therefore, not only the visual quality of camouflage image improved a lot by dividing  secret and target  image into smaller blocks for transformation, but also the reversibility is  achieved.

As for RVT, there are two steps to generate camouflage image, the  first step is dividing secret image
and  target image into  small blocks and transforming secret image to one target image to get a transformed image, and in the second step we embed some auxiliary parameters into the transformed image by RDH.  At the receiver's side, by the decode processes of RDH we restore the auxiliary parameters and the transformed image from the camouflage image firstly, and then the secret image is restored from the  transformed image with the help of the restored auxiliary parameters.  The diagram RVT is as Fig. \ref{RVTflow}.

\subsection{Reversible image processing}\label{subsec2.4}

Image processing is rather popular, and  people process their images to
desired results through various of tools \cite{rip0,rip1}. Nowadays, most of image processing methods are irreversible, that is to say after processing the image we will damage the original copy. However, sometimes the client may not  satisfy with the processed result, then the irreversibility will result in great inconvenience. Of course, the users can save the original
copy as a backup before processing it, which will cost much more storage space. Indeed, Google's  Picasa's automatic image enhancement system is one of such examples, who stores the original image in a separate folder as the backup.
Instead of storing both the original and the processed images, Apple and Google Photos keep the original image
and a small record file of the applied enhancements. The enhanced image will be displayed each time by re-enhanced the original image with the help of  the record file. However, the enhanced image can be only correctly viewed on their own software. What is more, processing original image for displaying each time will  waste many computing resources.

\begin{figure}[h!]
\centering
\includegraphics[width=0.45\textwidth]{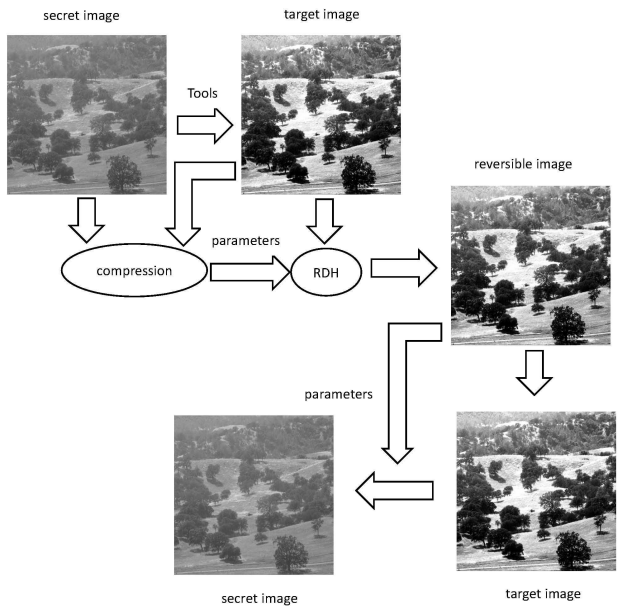}
\begin{center}
\caption{\small{Reversible image processing.}}\label{RIPflow}
\end{center}
\end{figure}

As shown in Fig. \ref{RIPflow}, technological process of reversible image processing \cite{rip} is described as follows. By using  algorithms
or softwares,  the original image is processed to  the
desired result regarded as target image.  Since the target image is obtained from the original image, the correlations between original image and  target image are high, thus can be explored to compress  the original image effectively. We get the reversible image similar to the target image by embedding the compressed secret image into the target image with an RDH scheme.

\begin{figure}[h!]
\centering
\includegraphics[width=0.45\textwidth]{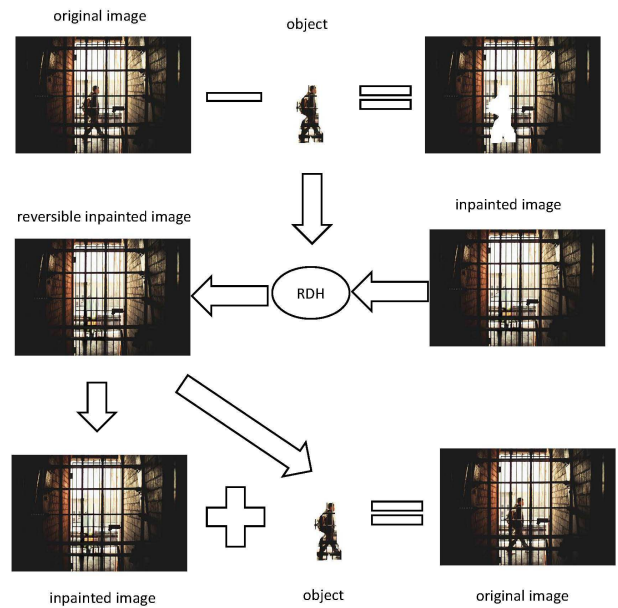}
\begin{center}
\caption{\small{Reversible image inpainting.}}\label{localRIP}
\end{center}
\end{figure}

Some  image processing algorithms only  precess local regions of one image, in such case, the amount of information for recording the processed region is  small, and it is much easy for reversibility. Taking inpainting as example shown in Fig. \ref{localRIP}, we cut out the person in original image, and then inpaint the remained content to make it natural. To make the operation reversible, we only need to reversibly embed the cropped person into the inpainted image to get the reversible inpainted image.

\section{Conclusion and discussion}\label{sec3}
In the past  years, the motivation of RDH is mainly about integrity
authentication. Now, we present some innovate applications based on RDH, containing reversible steganography, reversible image processing, reversible adversarial example, reversible visual transformation. Of course, besides those, we believe that style transfer \cite{st1,st2,st3,st4}, colorization\cite{color1,color2}, and so on, cam be also reversible.
These applications have  valuable prospects and extend the application of RDH a lot.

There are still many difficulties to be solved in these reversible operations. For example, compared to adversarial example the accuracy of reversible adversarial example will be slightly  descended. As for some complex image processing methods, the auxiliary information for restoring original image is too much to be embedded by RDH, or will greatly degrade the quality of the processed
image. In the future, we will try to overcome those difficulties to make these reversible operations be more better.

\section{Acknowledgments}\label{sec3}
This work was supported in part by the Natural Science
Foundation of China under Grant U1636201 and 61572452.

\begin{table*}[t!]
\caption{Authors' background.}
\centering
\begin{tabular}
{ p{55pt} p{45pt} p{125pt} p{165pt}   }
\hline
$Your Name$ & $Title$ & $Research Field$ & Personal website  \\
\hline
Dongdong Hou &   Phd candidate & data hiding and image processing& http://home.ustc.edu.cn/~houdd/ \\
\hline
Weiming Zhang &  Full professor & data hiding and image processing & http://staff.ustc.edu.cn/~zhangwm/index.html \\
\hline
Jiayang Liu &  Phd candidate & data hiding and image processing & ~ \\
\hline
Dongdong Chen &  Phd candidate & image processing & ~ \\
\hline
Siyan Zhou &  master student & image processing & ~ \\
\hline
Nenghai Yu &  Full professor & data hiding and image processing& http://staff.ustc.edu.cn/~ynh/ \\
\hline
\end{tabular}
\label{cor}
\end{table*}

\end{document}